\documentclass[conference]{IEEEtran}
\usepackage{cite}
\usepackage{amsmath,amssymb,amsfonts}
\usepackage{algorithmic}
\usepackage{graphicx}
\usepackage{textcomp}
\usepackage{xcolor}
\usepackage{cite}
\usepackage{colortbl}
\usepackage{multirow}
\usepackage{afterpage}
\usepackage{url}
\usepackage{pifont}
\usepackage{hyperref}
\usepackage{enumitem}

\def\BibTeX{{\rm B\kern-.05em{\sc i\kern-.025em b}\kern-.08em
    T\kern-.1667em\lower.7ex\hbox{E}\kern-.125emX}}
\begin{document}

\title{MedThink: Explaining Medical Visual Question Answering via Multimodal Decision-Making Rationale\\
 \thanks{Identify applicable funding agency here. If none, delete this.}
}

\author{
    \IEEEauthorblockN{
        Xiaotang Gai\textsuperscript{1}\textsuperscript{*}, 
        Chenyi Zhou\textsuperscript{1}\textsuperscript{*}, 
        Jiaxiang Liu\textsuperscript{1}\textsuperscript{*}, 
        Yang Feng\textsuperscript{2}, 
        Jian Wu\textsuperscript{1}, 
        Zuozhu Liu\textsuperscript{3}\textsuperscript{\#}
    }
    \IEEEauthorblockA{\textsuperscript{1}Zhejiang University, Zhejiang, China\\
    Email: \{xiaotang.23, chenyi.22, jiaxiang.21\}@intl.zju.edu.cn, wujian2000@zju.edu.cn}
    \IEEEauthorblockA{\textsuperscript{2}Angelalign Tech, Zhejiang, China\\
    Email: fengyang@angelalign.com}
    \IEEEauthorblockA{\textsuperscript{3}Zhejiang University, Zhejiang, China\\
    Email: zuozhuliu@intl.zju.edu.cn}
    \IEEEauthorblockA{\textsuperscript{*}Equal contribution.  \textsuperscript{\#} Correspondence author.}
}

\maketitle

\begin{abstract}
Medical Visual Question Answering (MedVQA), which offers language responses to image-based medical inquiries, represents a challenging task and significant advancement in healthcare. It assists medical experts to swiftly interpret medical images, thereby enabling faster and more accurate diagnoses. 
However, the model interpretability and transparency of existing MedVQA solutions are often limited, posing challenges in understanding their decision-making processes. 
To address this issue, we devise a semi-automated annotation process to streamline data preparation and build new benchmark MedVQA datasets R-RAD, R-SLAKE and R-Path. These datasets provide intermediate medical decision-making rationales generated by multimodal large language models and human annotations for question-answering pairs in existing MedVQA datasets, i.e., VQA-RAD, SLAKE and PathVQA. Moreover, we design a novel framework, MedThink, which finetunes lightweight pretrained generative models by incorporating medical decision-making rationales. MedThink includes three distinct strategies to generate decision outcomes and corresponding rationales, thereby clearly showcasing the medical decision-making process during reasoning. Our comprehensive experiments show that our method achieves an accuracy of 83.5\% on R-RAD, 86.3\% on R-SLAKE and 87.2\% on R-Path. These results significantly exceed those of existing state-of-the-art models with comparable parameters. Datasets and code will be released. 
\end{abstract}


\section{Introduction}
The Medical Visual Question Answering (MedVQA) task is designed to take medical images and specialized clinical queries as inputs, and provide accurate answers with texts.
Since the inception of the MedVQA challenge in 2018\cite{hasan2018overview}, there has been a significant surge in interest in exploring the capabilities of MedVQA \cite{Liujiaxiang2023Adapter}. 
Effective MedVQA not only holds the potential to enhance patient engagement, thereby alleviating patient stress, but also assists physicians in clinical diagnosis, thus conserving valuable medical resources and reducing the risk of misdiagnosis \cite{zhan2020medical}.

The challenges to resolve the MedVQA tasks are two-fold. On one hand, though there exist a wealth of datasets composed of medical images and text annotations \cite{porwal2018indian}, 
the decision-making process between the question and answer pairs are usually missing, impeding reliable evaluation of model interpretability. While some recent datasets already incorporated images, specialized medical queries, and answer texts ~\cite{lau2018dataset,liu2021slake}, the corresponding reasoning process to reach certain diagnostic decisions remain unclear, resulting in black-box and clinically inapplicable inference~\cite{lu2022learn,liu2023chatgpt}. 
A straightfoward solution is to integrate expert-level reasoning rationales in these datasets to unravel the underlying reasoning processes. However, manual annotation of such rationales is time-consuming and requires in-depth understanding of medical knowledge, while a fast and reliable rationale annotation framework is still missing \cite{liu2023deep}. 

On the other hand, models which can resolve the MedVQA tasks in a fast, accurate and interpretable manner is of high necessity in real-world applications. Current MedVQA methods often model this problem by retrieval and train MedVQA models with contrastive or classification objectives. For instance, Nguyen et al. \cite{nguyen2019overcoming} employed a combination of unsupervised convolutional denoising autoencoders and the meta-learning method to learn domain specific weight initialization of MedVQA model on external medical datasets. Moreover, Zhang et al. \cite{zhang2022contrastive} first implemented contrastive learning in the medical domain, presenting ConVIRT, a methodology that utilizes medical text-image contrastive loss for pretraining medical visual representations. Further, Liu et al. \cite{liu2021contrastive} proposed CPRD, a two-stage pre-training framework, leveraging representation distillation and contrastive learning to train visual encoder for MedVQA system on a large corpus of unlabeled radiological images. The recent PubMedCLIP 
model \cite{eslami-etal-2023-pubmedclip} pioneers the incorporation of the Contrastive Language-Image Pre-Training \cite{radford2021learning} into the MedVQA tasks by conducting pre-training. 

In contrast, the remarkable performance of large language models (LLMs) across various natural language processing (NLP) tasks has been extended to text question-answering in healthcare\cite{nori2023capabilities} 
Building upon this, multimodal large language models (MLLMs) \cite{openai2023gpt4vision,team2023gemini} 
accept both text and image inputs to generate responses, presenting a novel approach to tackling the MedVQA tasks. However, applying MLLMs directly to the MedVQA tasks in real medical scenarios is impractical due to their high operational costs and significant latency. 

In this paper, we aim to address the aforementioned challenges by providing new benchmark datasets and novel MedVQA solutions. We design a semi-automated annotation method that leverages the powerful inference capabilities of MLLMs to assist experts during annotation, significantly improving the efficiency. Through our method, we develop the R-RAD, R-SLAKE and R-Path datasets. These datasets provide the intermediate reasoning steps critical for medical decision-making, including necessary medical background knowledge and descriptions of medical images, which we term Medical Decision-Making Rationales (MDMRs). Moreover, we design a novel framework, MedThink, to finetune the pretrained generative models, specifically selecting the T5-base architecture~\cite{raffel2020exploring} as our base architecture due to its practicality in real-world applications. With only 223M parameters, the architecture adeptly performs generative tasks, balancing cost-effectiveness and practical value. By incorporating MDMRs into the training process, our model outputs not only decision outcomes but also corresponding rationales, thereby clearly showcasing the medical decision-making process during inference. Based on different inputs for MDMRs during training, we further propose three distinct generative modes: ``Explanation", ``Reasoning", and ``Two-Stage Reasoning", as shown in Figure~\ref{fig: overview}.

Extensive experimental results demonstrate that our method achieves an accuracy of 83.5\% on R-RAD, 86.3\% on R-SLAKE and 87.2\% on R-Path. These results represent significant enhancements over the existing state-of-the-art models with comparable parameters. Our contributions are as follows:
\setlength{\leftmargini}{1.3em}
\begin{itemize}
    \item We develop a semi-automated process for annotating MedVQA data with decision-making rationale. To the best of our knowledge, the R-RAD, R-SLAKE and R-Path datasets represent the first MedVQA benchmark datasets that encompass rationales for answers.
    \item We propose a lightweight framework, MedThink, with three answering strategies,  enabling faster and more accurate MedVQA with enhanced interpretability. 
    \item We conduct extensive experiments and ablations that demonstrate the usefulness of the R-RAD, R-SLAKE and R-Path datasets and superiority of MedThink.  

\end{itemize}

\section{Related Work}

\subsection{MedVQA}
\afterpage{
\begin{figure*}[t]
    \centering
    \includegraphics[width=1\linewidth]{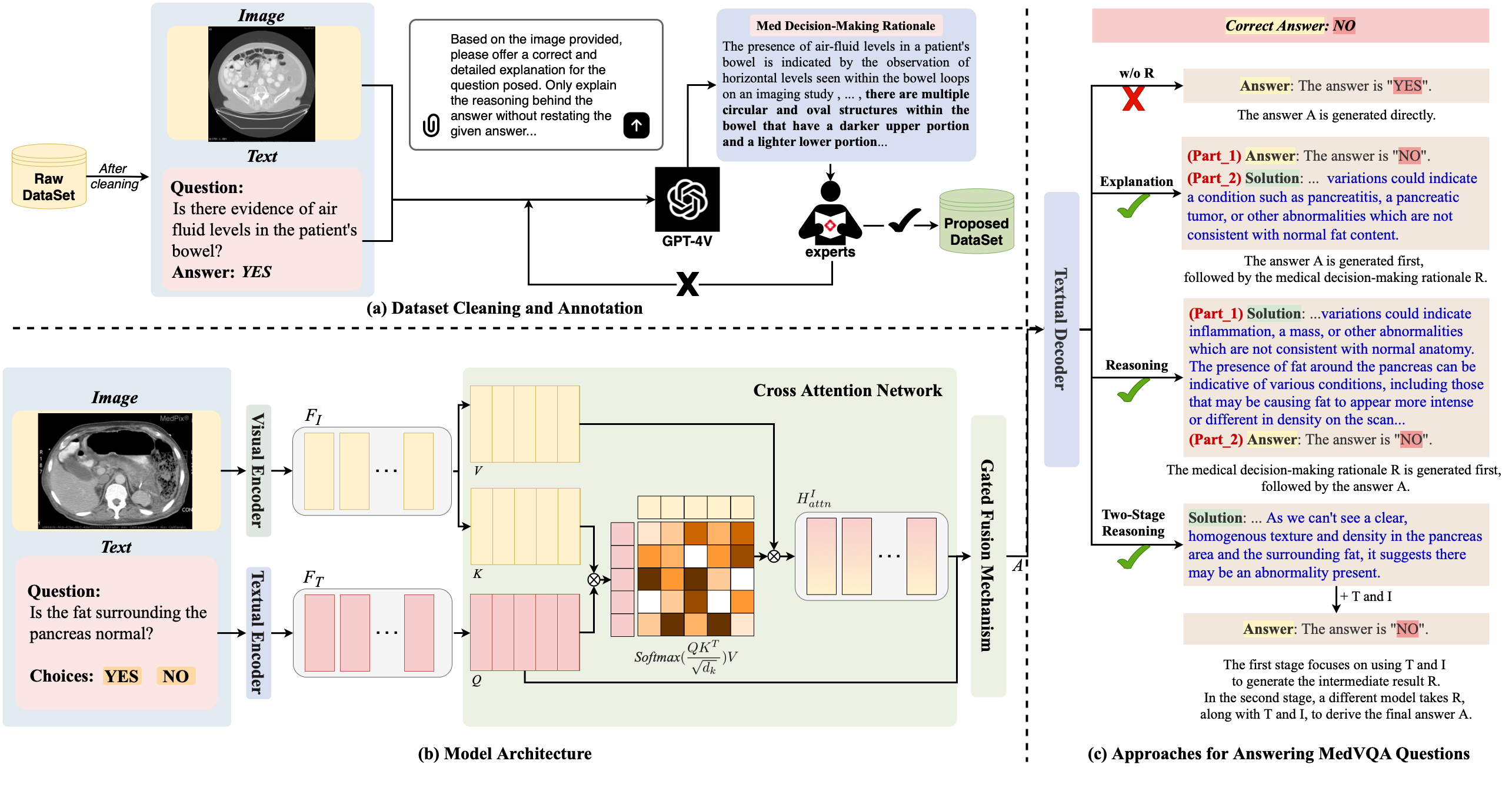}
    \caption{Overview of the Data Preparation, Model Architecture and Methods for Answering MedVQA Questions. \textbf{(a)} outlines the dataset cleaning and annotation process, where raw data undergoes refinement and annotation to formulate a new dataset with accurate MDMRs. \textbf{(b)} displays the model architecture, which incorporates a textual encoder for processing the medical question, 
    a visual encoder for analyzing medical images, and a cross-attention network with a gated fusion mechanism that synergistically combines textual and visual features to generate informed responses for the MedVQA task.~\cite{carion2020end, khashabi2020unifiedqa,zhang2023multimodal} \textbf{(c)} is the illustration of various strategies for answering MedVQA questions. These strategies show how the inclusion and arrangement of MDMRs can influence the model's output. The training process involves three steps. First, the training sets are annotated by the MLLM. Next, we use the training sets to train our models. Third, trained models generate MDMRs for the test sets.}
    \label{fig: overview}
    \vspace{-1em}
\end{figure*}

} 
VQA represents a cutting-edge, multimodal task at the intersection of computer vision and natural language processing, drawing significant attention in both domains. MedVQA applies the principles of VQA to interpret and respond to complex inquiries about medical imagery. A MedVQA system usually consists of three key components for feature extraction, feature fusion and answer reasoning, 
which aims to generate answers in text by processing given medical images. 

Previous MedVQA solutions  \cite{nguyen2019overcoming, zhang2022contrastive} 
have relied on the CNNs, such as those pretrained on ImageNet like VGGs or ResNets, to extract visual features. Meanwhile, the RNNs are employed to process textual information. With the development of large-scale pretraining, recent works \cite{Liujiaxiang2023Adapter,van2023open,eslami-etal-2023-pubmedclip} have shifted towards the transformer-based models to enhance feature extraction capabilities for both textual and visual modalities. 

In terms of content, these works still treat the MedVQA as the classification problem. However, this approach is misaligned with the realities of medical practice, where clinicians rarely face scenarios that can be addressed with predefined answer options.

This incongruity underscores the necessity for a MedVQA approach that is more adaptive and reflective of the complexities inherent in medical diagnostics and decision-making. Our paper redefines MedVQA as the generative task. Within actual medical environment, when faced with open-ended queries, our proposed MedVQA model can still generate informed responses based on the medical knowledge it has learned.
 
\subsection{The Chain of Thought }
Recently, NLP has been significantly transformed by language models \cite{raffel2020exploring,chowdhery2023palm}. 
To further enhance the reasoning capabilities of language models, prior works \cite{cobbe2021training,wei2022chain} have incorporated reasoning rationales during training or inference phases, which guide models to generate the final prediction. On the other hand, in the realm of VQA, it is crucial for VQA systems to understand multimodal information from diverse sources and reason about domain-specific questions. To achieve this goal, several works \cite{lu2022learn,zhang2023multimodal} have proposed multimodal reasoning methods for VQA. These methods, commonly referred to as ``Chain of Thought", introduces intermediate steps to assist the model in reasoning. In this paper, we present the ``Medical Decision-Making Rationale" (MDMR) and apply it to the MedVQA tasks. We anticipate that MedVQA systems, equipped with the MDMR, will not only offer support in medical decision-making but also elucidate the underlying rationales behind these decisions.

\section{Methodology}
\subsection{Problem Formulation}
In this paper, We denote the medical dataset as $\mathcal{D}=\{(I_m, T_m, A_m, R_m)\}^M_{m=1}$ where $M$ is the number of data samples. And the goal of the MedVQA tasks is to develop a mapping function $f(\cdot)$ that can generate textual answers in response to the medical questions, represented as:
\begin{equation}
    \{A, R\} = f(I, T),
\end{equation}
Here, $I$ denotes the medical image sourced from modalities such as X-ray, CT, or MRI. $T$ represents the natural language question pertaining to the medical image $I$. The output of the model $f(\cdot)$, represented as $\{A, R\}$, comprises two components. $A$ is the predicted textual answer, directly addressing the query posed in $T$. $R$, termed as ``medical decision-making rationale", offers a detailed justification for the answer $A$, elucidating an interpretative insight into how the model processes $I$ and $T$.

\subsection{Model Architecture}
The model architecture comprises five components, shown in Figure~\ref{fig: overview} (b): TextualEncoder, VisualEncoder, Cross Attention Network, Gated Fusion Network, and TextualDecoder. Notably, the TextualEncoder, VisualEncoder and TextualDecoder are all based on the Transformer architecture, renowned for its powerful learning and representational capabilities.

The TextualEncoder vectorizes the input question $T$ into the textual feature space, represented as $F_T \in\mathbb{R}^{n \times d}$, while the VisualEncoder transforms the input medical image $I$ into vision features $F_I \in\mathbb{R}^{m \times d}$. This can be expressed as: \( F_{\text{T}} = \text{TextualEncoder}(T) \) and \( F_{\text{I}} = \text{VisualEncoder}(I) \),
where $n$ denotes the length of the input text, and $d$ indicates the hidden dimension, $m$ represents the number of image patches.

Upon acquiring the textual representation $F_T$ and visual representation $F_I$, our model employs the 
Cross-Attention Network to facilitate interaction between these two modalities. The Cross-Attention Network computes the attention-guided visual feature $H_{attn}^I \in\mathbb{R}^{n \times d}$, which captures the relevant visual features corresponding to the textual query through the following operation:
\begin{align}
        H_{\text{attn}}^{\text{I}} &= \text{Softmax}\left(\frac{QK^T}{\sqrt{d}}\right)V,
\end{align}
where $Q$, $K$, $V$ correspond to the query, key, and value, derived from $F_T$, $F_I$, $F_I$, respectively.

Subsequently, the Gated Fusion Mechanism is utilized to dynamically combine the textual representation $F_T$ and the attention-guided visual feature $H_{attn}^I$. It determines the fusion coefficient $\lambda$ through a sigmoid-activated linear combination of the two modalities:
\begin{align}
         \lambda &= \text{Sigmoid}(W_lF_Q + W_vH_{\text{attn}}^{\text{I}}),
\end{align}

The fused output $F_{\text{fuse}} \in\mathbb{R}^{n \times d}$ is then computed as a weighted sum of $F_T$ and $H_{\text{attn}}^{\text{I}}$, moderated by $\lambda$:
\begin{align}
         F_{\text{fuse}} &= (1-\lambda) \cdot F_{\text{T}} + \lambda \cdot H_{\text{attn}}^{\text{I}},
\end{align}
Here, $W_l$ and $W_v$ are the model parameters that are learned during training to optimize the fusion of information between textual and visual streams.
Finally, the fused output $F_{\text{fuse}}$ is fed into the TextualDecoder to generate the output $\{A, R\}$:
\begin{align}
    \{A, R\} &= \text{TextualDecoder}(F_{\text{fuse}}), 
\end{align}

\begin{table}
  \centering
    \caption{Details of Datasets: Distribution of Images and Questions in the R-RAD, R-SLAKE and R-Path Datasets.}
  \scalebox{1.1}{
  \begin{tabular}{cccc}
    \hline
    \textbf{Dataset}      & \textbf{Images} & \textbf{Training set} & \textbf{Test set} \\\hline
    R-RAD (closed-end)    & 300             & 1823                  & 272      \\
    R-RAD(open-end)       & 267             & 1241                  & 179      \\\hline
    R-SLAKE(closed-end)   & 545             & 1943                  & 416      \\
    R-SLAKE(open-end)     & 545             & 2976                  & 645      \\\hline
    R-PathVQA(closed-end) & 3361            & 9806                  & 3391     \\
    R-PathVQA(open-end)   & 3425            & 9933                  & 3364     \\\hline
  \end{tabular}
  }
  \label{tab: dataset}
  \vspace{-1.5em}
\end{table}
\subsection{Loss Function}
Given the input $X=\{I,T\}$, the model $f$ is trained by maximizing the likelihood of accurately predicting the target output $Y$ = $\{A, R\}$. The training involves a loss function, primarily the negative log-likelihood of correctly predicting subsequent tokens in the sequence $Y$, accumulated over all time steps. This is mathematically formulated as:
\begin{equation}
    L = -\sum_{n=1}^{N}\log p(Y_n|X, Y^{1:n-1}),
\end{equation}
In this context, $N$ represents the total number of tokens in the target answer $Y$, and $p(Y_n|X, Y^{1:n-1})$ denotes the conditional probability of correctly predicting the $n$-th token in $Y$, given the input $X$ and all preceding tokens $Y^{1:n-1}$ in the sequence. This loss function significantly improves the model's capability to accurately forecast each token in the target output, thereby enhancing its overall predictive performance.

\subsection{Three Generation Strategies}
To investigate the impact of MDMRs on the model performance in the MedVQA tasks, we present three different generation strategies. These strategies are designed to guide the model in generating various forms of outputs, corresponding to different orders of MDMR in the process of generation. The methods are categorized as ``Explanation", ``Reasoning" and ``Two-Stage Reasoning", as shown in Figure~\ref{fig: overview} (c).

In the “Explanation” method, the answer $A$ is generated first, followed by the MDMR $R$. In contrast, the “Reasoning” method reverses this order, generating $R$ before $A$. The ``Two-Stage Reasoning" method follows a phased strategy, where two independent models are trained in distinct stages. The first stage focuses on using the medical question $T$ and the medical image $I$ to generate the intermediate result $R$. In the second stage, a different model takes $R$, along with $T$ and $I$, to derive the final answer $A$.
\begin{table*}
\centering
\caption{Accuracy (\%) Comparison of Methods on Closed-End Questions in the R-RAD, R-SLAKE and R-Path Datasets.}
\scalebox{1.16}{
\begin{tabular}{ccccccc}
\hline
 Methods &MLLM-Based  &   \textbf{R-RAD} &  \textbf{R-SLAKE} &  \textbf{R-Path}  & Parameters\\
 
\hline
\multicolumn{6}{c}{\textit{Zero-shot results}} \\
\hline
\rowcolor{gray!10}  Med-MoE(StableLM)\cite{jiang2024medmoemixturedomainspecificexperts} &\ding{51}  &  66.9 & 52.6 & 69.1  & 2B \\
\rowcolor{gray!10}   LLaVA-Med(From LLaVA) \cite{li2024llava} &\ding{51}  &  60.2 & 47.6 & 59.8  &7B \\
\rowcolor{gray!10}  Gemini Pro\cite{team2023gemini}   &\ding{51} & 73.5 & 69.0 & 64.8  & - \\
\rowcolor{gray!10}  Gemini Pro (w/ Reasoning)\cite{team2023gemini} &\ding{51}  &  77.2 & 77.4 & 70.9  & - \\
\rowcolor{gray!10}  Gemini Pro (w/ Two-Stage Reasoning)\cite{team2023gemini} &\ding{51}  &  79.4 & 77.9 & 72.3 & - \\
\rowcolor{gray!10}  Gemini Pro (w/ Explanation)\cite{team2023gemini} & \ding{51}  &  79.8 & 78.1 & 72.6  & - \\
\hline
\multicolumn{6}{c}{\textit{Representative \& SOTA methods (Supervised finetuning results)}} \\
\hline
\rowcolor{gray!10}     MFB\cite{yu2017multi} & & 74.3 & 75.0 & -  &-  \\
\rowcolor{gray!10}  SAN\cite{yang2016stacked} &  & 69.5 & 79.1 & -  &-  \\
\rowcolor{gray!10}   BAN\cite{kim2018bilinear} & & 72.1 & 79.1 & -  &-  \\
\rowcolor{gray!10}  MEVF+SAN\cite{nguyen2019overcoming}& &  73.9 & 78.4 & -  &- \\
\rowcolor{gray!10} MEVF+BAN\cite{nguyen2019overcoming} &  & 77.2 & 79.8 & -  &-  \\
\rowcolor{gray!10} MMBERT\cite{tiong-etal-2022-plug} & &-& 77.9 & - & -\\
\rowcolor{gray!10} PubMedCLIP\cite{eslami-etal-2023-pubmedclip} &  & 79.5 & 82.5 & -  &-  \\
\rowcolor{gray!10} Prefix T. Medical LM(GPT2-XL)\cite{van2023open}  & \ding{51}  & - & 82.1 & 87.0  &1.5B \\
\rowcolor{gray!10} LLaVA~\cite{li2024llava} & \ding{51}  & 65.1 & 63.2 & 63.2  &7B \\
\rowcolor{gray!10} Med-Flamingo~\cite{moor2023med} & \ding{51}  & 65.1 & 63.2 & 63.2  &7B \\
\rowcolor{gray!10} LLaVA-Med (From LLaVA)~\cite{li2024llava} &\ding{51}  & 84.2 & 85.3 & 91.2  &7B \\
\rowcolor{gray!10} LLaVA-Med (From Vicuna)~\cite{li2024llava} &\ding{51}  & 82.0 & 83.2 & 91.7  &7B  \\
\rowcolor{gray!10} Med-MoE(StableLM)~\cite{jiang2024medmoemixturedomainspecificexperts} & \ding{51}  & 80.1 & 83.4 & 91.3  &2B \\
\rowcolor{gray!10} Med-Gemini~\cite{yang2024advancing} & \ding{51}  & - & 84.8 & 83.3  & - \\
\rowcolor{red!10} \textbf{MedThink (w/o R)} &  & \textbf{79.0} & \textbf{82.5} & \textbf{86.0}  &\textbf{0.2B} \\
\rowcolor{red!10} \textbf{MedThink (w/ Reasoning)} &  & \textbf{73.9} \textcolor[rgb]{0,0,1}{(-5.1)} & \textbf{80.8} \textcolor[rgb]{0,0,1}{(-1.7)} & \textbf{83.1} \textcolor[rgb]{0,0,1}{(-2.9)}  &\textbf{0.2B} \\
\rowcolor{red!10} \textbf{MedThink (w/ Two-Stage Reasoning)} &  & \textbf{80.5} \textcolor[rgb]{1,0,0}{(+1.5)} & \textbf{79.1} \textcolor[rgb]{0,0,1}{(-3.4)} & \textbf{87.2} \textbf{\textcolor[rgb]{1,0,0}{(+1.2)}}  &\textbf{0.2B} \\
\rowcolor{red!10} \textbf{MedThink (w/ Explanation)} &  & \textbf{83.5} \textbf{\textcolor[rgb]{1,0,0}{(+4.5)}} & \textbf{86.3} \textbf{\textcolor[rgb]{1,0,0}{(+3.8)}} & \textbf{87.0} \textcolor[rgb]{1,0,0}{(+1.0)}  &\textbf{0.2B} \\
\hline
\multicolumn{6}{c}{\textit{\textsuperscript{*}Red and blue numbers indicate increases and decreases in accuracy compared to the \textbf{MedThink (w/o R)} results respectively.}}\\
\hline
\end{tabular}
}
\label{table:main}
 \vspace{-1.3em}
\end{table*}
\section{Dataset Creation}
\subsection{Dataset Collection}
We establish three benchmark datasets R-RAD, R-SLAKE and R-Path based on the VQA-RAD dataset~\cite{lau2018dataset}, the SLAKE dataset~\cite{liu2021slake} and the PathVQA~\cite{he2020pathvqa}, respectively. 

The VQA-RAD dataset sources its radiographic images from MedPix\textregistered, an open-access radiology database. In this dataset, clinicians formulate perinent medical questions based on the radiographic images, and provide corresponding answers. The VQA-RAD dataset comprises a collection of 315 images and 3,515 questions-answer pairs.

The SLAKE dataset derives its data from three distinct sources: the ChestX-ray8 \cite{wang2017chestx}, the CHAOS Challenge \cite{kavur2021chaos}, and the Medical Segmentation Decathlon (MSD) \cite{simpson2019large}. After screening and annotation by clinicians, it yields a bilingual (English-Chinese) MedVQA dataset, including 642 medical images and approximately 14,000 medical questions. For our work, we utilize only the ``English" component of the dataset.

The PathVQA dataset, specifically designed for visual question answering in the medical domain, compiles its pathology images and corresponding captions from a range of textbooks and online digital libraries. 
The dataset consists of 4,289 pathology images and 32,632 question-answer pairs, each pair is meticulously reviewed for accuracy. 

These questions of three datasets are classified as ``closed-end" if they have limited answer choices, and ``open-end" otherwise. For our work, We adhere to the official dataset split for evaluation. After completing the data cleaning and annotation, the R-RAD dataset includes a total of 3,515 medical questions and 314 medical images, the R-SLAKE dataset comprises 5,980 medical questions and 546 medical images and the R-Path dataset contains 4,012 images and 26,494 question-answer pairs. Relevant statistics for the R-RAD, R-SLAKE  and R-Path datasets are detailed in Table~\ref{tab: dataset}.

\subsection{Dataset Cleaning}

We identify noticeable inconsistencies within raw datasets. Specifically, the answers to similar questions about the same medical image are not always consistent. For instance, given a chest X-ray imaging, the response to the question ``Is/Are the right hemidiaphragm normal?" is ``No", while the answer to ``Is this image normal?" is ``Yes". This apparent contradiction prompted us to seek further expert medical review for such cases, ensuring the reliability of our dataset.

In light of advancements of MLLMs, we integrate the MLLM into our data cleaning and annotation process, aiming to streamline the workflows. This integration can not only expedite data processing but also unearth subtleties often missed in manual cleaning and annotation practices. To address inconsistencies, we first use the MLLM to systematically review all question-answer pairs for each medical image. After identifying inconsistencies, domain experts revise the answers, ensuring consistency across all questions related to the same medical image.

\subsection{Dataset Annotation}
After data cleaning, we utilize the MLLM for data annotation, specifically in generating MDMRs for the items within the VQA-RAD, SLAKE and PathVQA datasets, as shown in Figure~\ref{fig: overview} (a). This involves furnishing the MLLM with the datasets' images, questions, and correct answers. 
Therefore, we design a fixed prompt to guide the generation process of the MLLM. 
To ensure the quality of MDMRs, domain experts check MDMRs' validity and applicability. MDMRs not meeting criteria will be regenerated by the MLLM. If a MDMR generated by the MLLM remains below standard even after three attempts, domain experts will personally create an acceptable version, adhering to predefined criteria.

We enlist experienced physicians as domain experts to ensure the professional and accurate annotation of our data. To account for the diversity of medical opinions, we establish rigorous review criteria to guide the annotation process. The criteria are as follows:

(1) Coherence: The MDMR must be logically coherent, with no errors in grammar or spelling.

(2) Relevance: The MDMR must be directly related to the question and pertinent to the clinical context.

(3) Accuracy: The MDMR should be free from common sense and medical knowledge errors.

Only when all three conditions are met will the MDMR be included in our datasets.

\vspace*{-0.2cm}
\section{Experiments}
\subsection{Training Details}

During the datasets construction phase, we select GPT-4V~\cite{openai2023gpt4vision} from among MLLMs to handle data cleaning and annotation.
In our framework, the encoder and decoder from UnifiedQA \cite{khashabi2020unifiedqa} are integrated as TextualEncoder($\cdot$) and TextualDecoder($\cdot$), respectively. Additionally, DETR \cite{carion2020end} is employed as VisualEncoder($\cdot$).

In our experiments, the learning rate is uniformly set at 5e-4 for the R-SLAKE, R-RAD and R-Path datasets. The number of epochs during fine-tuning varies by dataset: 300 epochs for the R-SLAKE dataset, 150 epochs for the R-RAD dataset and 50 epochs for the R-Path dataset. It is important to note that our ``Two-Stage Reasoning" strategy requires a phased fine-tuning process involving two separate models. In the first phase, we follow the parameters mentioned above. In the second phase, we fine-tune with a learning rate of 5e-5 for 20 epochs across all three datasets. The batch size is set to 32.

\begin{table*}
  \centering
    \caption{Score (\%) Comparison of Medthink on Open-End Questions in the R-RAD, R-SLAKE and R-Path Datasets.}
  \scalebox{1.2}{
  \begin{tabular}{ccccccccc}
    \hline
 \textbf{Dataset} & \textbf{Strategy}         & \textbf{Rouge-1} & \textbf{Rouge-2} & \textbf{Rouge-L} & \textbf{BLEU-1} & \textbf{BLEU-2} & \textbf{BLEU-3} & \textbf{BLEU-4} \\\hline
  \multirow{3}{*}{R-RAD}   
        & Reasoning           & 49.8   & \textbf{20.3}   & 29.3   & 37.8   & 22.7   & \textbf{14.0}   & \textbf{8.9}    \\
        & Two-Stage Reasoning & 49.1   & 19.9   & 28.7   & 37.7   & 22.5   & 13.9   & 8.8    \\
        & Explanation         & \textbf{50.2}   & 20.2   & \textbf{29.5}   & \textbf{38.3}   & \textbf{22.9}   & \textbf{14.0}   & 8.8    \\
        \hline

\multirow{3}{*}{R-SLAKE} 
        & Reasoning           & \textbf{53.5}   & 22.8   &\textbf{32.1}   & \textbf{39.5}   & 24.3   & 15.5   & 10.0   \\
        & Two-Stage Reasoning & 53.2   & \textbf{23.1}   & 32.0   & \textbf{39.5}   & \textbf{24.5}   & \textbf{15.8}   & \textbf{10.3} \\
        & Explanation         & 53.1   & 22.7   & 31.7   & 39.2   & 24.1   & 15.4   & 9.9    \\
        \hline  
\multirow{3}{*}{R-Path} 
        & Reasoning & 41.5 & 13.0  & 24.8 & 31.8 & 17.0  & 9.6  & 5.7   \\
        & Two-Stage Reasoning & 41.7 & \textbf{13.2} & 24.9  &  \textbf{32.1}  &  \textbf{17.1}  & \textbf{9.7}   & \textbf{5.8} \\
        & Explanation & \textbf{41.9}  &  \textbf{13.2}  & \textbf{25.0}  & \textbf{32.1}   & \textbf{17.1}  & \textbf{9.7}   & \textbf{5.8}    \\
        \hline  
  \end{tabular}
  }
  \label{tab: open-end acc}
   \vspace{-1.7em}
\end{table*}

All experiments reported in this paper are conducted using PyTorch on an Ubuntu server equipped with four NVIDIA RTX 3090 GPUs. Training on the R-RAD dataset takes about 2.5 hours. In comparison, training on the R-SLAKE dataset requires approximately 5.5 hours, while the R-Path dataset takes around 14 hours. During inference, processing each sample takes about 6 seconds. 

\subsection{Evaluation Metrics}
Our performance evaluation is divided into two parts, focusing on closed-end and open-end questions separately. For closed-end questions, which are formatted as multiple-choice with a single correct answer, we assess performance using accuracy as the metric. 
For open-end questions, in contrast to previous works~\cite{yang2016stacked, kim2018bilinear, yu2017multi, nguyen2019overcoming, tiong-etal-2022-plug, eslami-etal-2023-pubmedclip} that often emphasize scoring all possible answers in open-ended MedVQA datasets to gauge classification accuracy, our work on generative MedVQA prioritizes clinical utility. Following established studies~\cite{li2023huatuo, zhang2023huatuogpt}, we employ BLEU and ROUGE to assess the quality of our method's outputs. The BLEU scores, akin to the ``Precision", evaluates the overlap of k-grams between generated and reference sentences, while the ROUGE scores, similar to the ``Recall", measures the similarity in word sequences.

\afterpage{
\begin{figure}[t]
    \centering
    \includegraphics[width=0.96\linewidth]{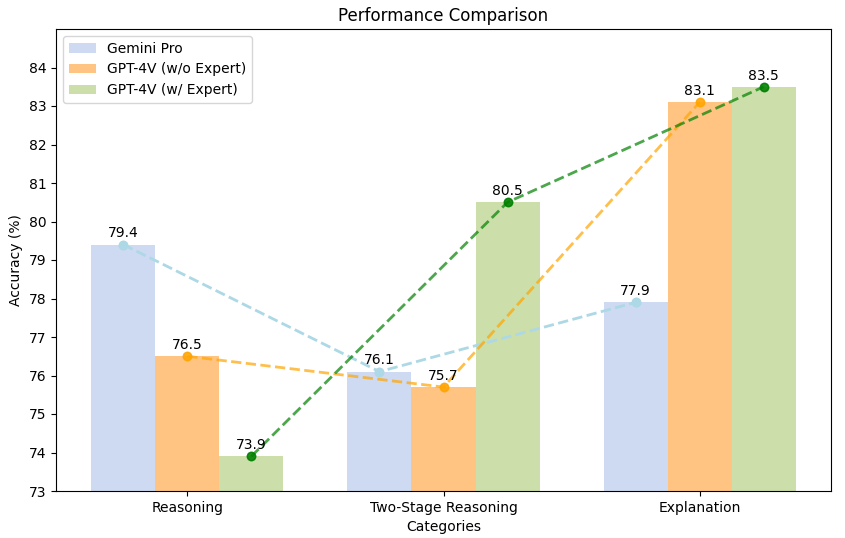}
    \caption{Impact of MLLMs Selection and Expert Participation in Dataset Creation on the MedVQA Tasks Accuracy (\%).}
    \label{fig: expert}
    \vspace{-1em}
\end{figure}

}
\subsection{Main Results}

In facing closed-end questions, we evaluate the performance of MedThink under various generation strategies, and compare them against several baseline methods on the R-RAD, R-SLAKE, and R-Path datasets. The results are shown in Table~\ref{table:main}.
MedThink demonstrates varying levels of performance across different generation strategies. To be specific, MedThink with the ``Explanation" strategy achieves the highest accuracy on the R-RAD and R-SLAKE datasets, recording 83.5\% and 86.3\% respectively. Meanwhile, MedThink with the ``Two-Stage Reasoning" strategy achieves the best performance on R-Path with an accuracy of 87.2\%. 

In contrast, the state-of-the-art classification model, PubMedCLIP, achieves accuracies of 79.5\% on the R-RAD dataset and 82.5\% on the R-SLAKE dataset, which is significantly lower than MedThink's results. This underscores the superior performance of MedThink. Compared to other generative models based on MLLM, MedThink outperforms models such as LLaVA~\cite{li2024llava}, Med-Flamingo~\cite{moor2023med}, and Med-Gemini~\cite{yang2024advancing}, achieving overall accuracies that are on par with the more parameter-heavy LLaVA-Med~\cite{li2024llava} and Med-MoE~\cite{jiang2024medmoemixturedomainspecificexperts} models. Notably, MedThink accomplishes this with a parameter count that is less than one-tenth of these models, demonstrating both its efficiency and effectiveness.

Using open-end questions, we conduct a comprehensive evaluation of MedThink's three strategies on the R-RAD, R-SLAKE, and R-Path datasets. The results are summarized in Table~\ref{tab: open-end acc}.
For the R-RAD dataset, the "Explanation" strategy outperforms other strategies, achieving the highest scores in five out of seven metrics. It records 50.2\% in Rouge-1, 29.5\% in Rouge-L, 38.3\% in BLEU-1, 22.9\% in BLEU-2, and 14.0\% in BLEU-3. On the R-SLAKE dataset, the "Two-Stage Reasoning" strategy leads in performance, securing the top scores in five out of seven metrics, with 23.1\% in Rouge-2, 39.5\% in BLEU-1, 24.5\% in BLEU-2, 15.8\% in BLEU-3, and 10.3\% in BLEU-4. Regarding the R-Path dataset, the "Explanation" strategy once again delivers the highest overall performance, achieving 41.9\% in Rouge-1, 13.2\% in Rouge-2, 25.0\% in Rouge-L, 32.1\% in BLEU-1, 17.1\% in BLEU-2, 9.7\% in BLEU-3, and 5.8\% in BLEU-4. These results collectively highlight the importance of selecting appropriate generation strategies tailored to addressing various medical scenarios, ensuring the model generates comprehensive and detailed responses.

\afterpage{
\begin{figure}[t]
    \centering
    \includegraphics[width=1\linewidth]{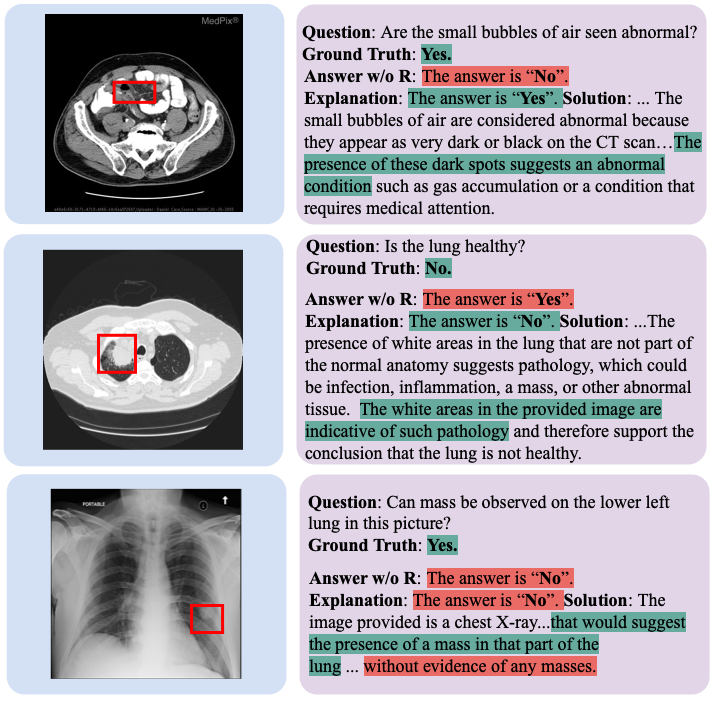}
    \caption{Illustration of MDMRs Enhancing Model Responses in the MedVQA Tasks. The green highlighted text represents medically relevant knowledge that aids in answering the question, while the red highlighted text indicates information that could lead to incorrect conclusions. The red boxes in the images correspond to the described anatomical features, underscoring the alignment between the rationale and the visual evidence.}
    \label{fig: case study}
\vspace{-1em}
\end{figure}
}
\subsection{Ablation Study}
To explore the influence of various components in MedThink, we conduct a series of ablation experiments.
First, we evaluate the effect of different MLLMs used during dataset creation and the contribution of domain experts to data annotation. We implement three variations for annotating the closed-end questions in the R-RAD dataset: using Gemini Pro~\cite{team2023gemini} without expert involvement, using GPT-4V without expert involvement, and using GPT-4V with expert involvement. As shown in Figure~\ref{fig: expert}, when GPT-4V is used with expert involvement in dataset creation, the ``Explanation" and ``Two-Stage Reasoning" strategies achieved their highest accuracies of 83.5\% and 80.5\%, respectively. In contrast, the 'Reasoning' strategy performed best with Gemini Pro without expert involvement, reaching an accuracy of 79.4\%, which is only slightly above the baseline accuracy of 79.0\% when MDMRs are not applied. We attribute this to the instability of the 'Reasoning' strategy, which hinders its ability to consistently benefit from MDMRs, aligning with previous research~\cite{lu2022learn}. Overall, expert involvement enhances the quality of MDMRs, positively impacting MedThink. Additionally, GPT-4V's stronger reasoning ability compared to Gemini Pro~\cite{fu2023challenger} further suggests that using a more advanced MLLM during data annotation is beneficial.

Next, we examine how the introducing of MDMRs impacted results. We introduce a control experiment, MedThink without MDMRs, where the models are trained and inferred without incorporating MDMRs (MedThink w/o R). The ``Explanation", ``Reasoning" and ``Two-Stage Reasoning" strategies are compared with the control experiment. As indicated in Table~\ref{table:main}, compared to ``MedThink w/o R", the ``Explanation", ``Two-Stage Reasoning", and ``Reasoning" strategies improve accuracy by 4.5\%, 1.5\%, and -5.1\% on the R-RAD dataset, 3.8\%, -3.4\%, and -1.7\% on the R-SLAKE dataset, and 1.0\%, 1.2\% and -2.9\% on the R-Path dataset, respectively. 

Finally, we assess the practicality of MDMRs generated by MedThink using different strategies. Initially, Gemini Pro is provided with only medical queries and related imagery. Subsequently, we inincorporate MDMRs generated by MedThink with the ``Explanation", ``Reasoning" and ``Two-Stage Reasoning" strategies to assist Gemini Pro in answering. The results, presented in Table~\ref{table:main}, indicate an initial accuracy of 73.5\% on the R-RAD dataset, 69.0\% on the R-SLAKE dataset and 64.8\% on the R-Path dataset for Gemini Pro. The integration of MDMRs has led to significant improvements. Among three strategies, the ``Explanation" strategy stands out, enhancing the accuracy by 6.3\% on the R-RAD dataset, 9.1\% on the R-SLAKE dataset and 7.8\% on the R-Path dataset.

\subsection{Case Study}
To assess the specific impact of MDMRs on the MedVQA task, Figure~\ref{fig: case study} shows several examples where MedThink applies the ``Explanation" strategy to answer questions from the R-SLAKE datasets. 

When the generated MDMR is accurate, MedThink can effectively and precisely answer the related medical question. If the MDMR contains errors, however, it misguides MedThink, leading to a phenomenon known as hallucination, which is a common issue in vision-language models. To investigate the causes of hallucinations in MedThink, we analyze the number of incorrect answers it provided on the R-SLAKE dataset. The R-SLAKE dataset is chosen because it covers medical questions about six anatomical regions, offering a complex and representative challenge.

We perform the analysis through the following steps. First, we categorize the test set questions by the anatomical regions associated with the medical images. Next, we tally the number of incorrectly predicted questions for each anatomical region. Finally, we calculate the proportion of incorrect predictions for each region, as shown in Table~\ref{table: analysis}. The results indicate that MedThink significantly aids in addressing medical issues related to the chest and abdomen. However, these areas still account for the majority of prediction errors. We attribute this to the greater complexity of chest and abdomen images, which contain more organs than other regions, presenting a considerable challenge for the model.

\section{Conclusion}
In this paper, we present a generative model-based framework for MedVQA and construct the R-RAD, R-SLAKE and R-Path datasets, which include intermediate reasoning steps to address the challenge of black-box decision-making processes in MedVQA models. Extensive experimental results show that our proposed framework not only elucidates the medical decision-making process of MedVQA models with clarity but also significantly enhances their performance. Future research will further explore generative models tailored for real clinical settings and how to better evaluate the performance of MedVQA models in open-ended scenarios. 

\begin{table}[t]
\centering
\caption{The error rate for each region (Lower values are better)}
\scalebox{1.05}{
\begin{tabular}{ccc}
\hline 
\textbf{Anatomical Regions (Number)}   & \textbf{w/o Rationale} $\downarrow$ & \textbf{Explanation} $\downarrow$ \\
\hline
Lung (N=141) & 12.06\% & 9.93\% \\
Abdomen (N=141) & 24.11\% & 19.15\% \\
Head (N=91) & 18.68\% & 13.18\% \\
Neck (N=16) & 18.75\% & 6.25\% \\
Chest (N=5) & 20.00\% & 0.00\% \\
Pelvic Cavity (N=22) & 4.55\% & 13.64\% \\
\hline
\end{tabular}
}
\label{table: analysis}
\end{table}
\bibliographystyle{IEEEtran}
\bibliography{custom}

\end{document}